\documentclass{article} 
\usepackage{nips14submit_e,times}
\usepackage{hyperref}
\usepackage{url}
\usepackage{epsfig}
\usepackage{graphicx}
\usepackage{amsmath}
\usepackage{amssymb}
\usepackage{booktabs}
\usepackage{multirow}
\usepackage{subfigure}
\usepackage{lib}

\def\loss{\textrm{loss}}

\title{R-CNNs for Pose Estimation and Action Detection}

\author{
Georgia Gkioxari \\
\small{University of California, Berkeley}\\
\texttt{gkioxari@berkeley.edu} 
\And
Bharath Hariharan \\
\small{University of California, Berkeley}\\
\texttt{bharath2@eecs.berkeley.edu} 
\And
Ross Girshick \\
\small{University of California, Berkeley}\\
\texttt{rbg@berkeley.edu}
\And
Jitendra Malik \\
\small{University of California, Berkeley}\\
\texttt{malik@berkeley.edu} 
}

\nipsfinalcopy 

\begin{document}

\maketitle

\begin{abstract}
We present convolutional neural networks for the tasks of keypoint (pose) prediction and action classification of people in unconstrained images. Our approach involves training an R-CNN detector with loss functions depending on the task being tackled. 
We evaluate our method on the challenging PASCAL VOC dataset and compare it to previous leading approaches. Our method gives state-of-the-art results for keypoint and action prediction. Additionally, we introduce a new dataset for \emph{action detection}, the task of simultaneously localizing people and classifying their actions, and present results using our approach.
\end{abstract}
\section{Introduction}
\seclabel{intro}
In this paper we investigate a deep learning approach to the problems of pose estimation and action classification. We build on the R-CNN \cite{girshick2014rcnn} object detection framework by training convolutional neural networks (CNNs) for each task.

R-CNN is a \emph{monolithic} approach, since it trains a one component detector. This is different from previous state-of-the-art methods in object detection, such as DPM \cite{lsvm-pami}, where multiple components and latent part models are used, or poselets \cite{BourdevPoseletsECCV10,kposelets}, where an ensemble of supervised part detectors are combined to make better object and keypoint predictions. 

Like R-CNN, our method takes as input an object proposal.
However, in addition to predicting the object class (person versus not person), our method also makes a confidence-scored prediction for the person's pose and action. For the task of pose estimation, our system outputs a set of keypoints, with scores, for each object proposal.
These predictions are evaluated under a detection setting, by measuring the AP for each keypoint independently \cite{yang2012articulated}. Our approach achieves a mean AP of 15.2\% on the PASCAL VOC 2009 person detection val dataset\footnote{This is the standard dataset for benchmarking human pose estimation in PASCAL \cite{armlets2013}}, which is an improvement over the previous state-of-the-art \cite{kposelets} mean AP of 12.7\%. For the task of action classification, for each proposal a score is associated with the action prediction. The evaluation measures AP for each action independently. Our approach achieves 70.5\% mAP on the PASCAL VOC action test set for action classification and is slightly better than the previous leading method \cite{Oquab14} (70.2\%), which also uses CNNs.

The standard method for evaluating action classification (reported above) assumes that ground-truth object locations are given at test time and one only needs to output an action label.
Knowledge of the ground truth at test time makes this task unrealistic for real-world applications. Therefore, we introduce a new task, which we call \emph{action detection}. In action detection, ground truth is unavailable at test time, thus an approach is expected to make predictions of the location of the person and the action being performed. Evaluation follows the standard AP metric. 
Intuitively, one can think of this task as introducing new object categories defined by (person, action) pairs, and then applying the standard PASCAL VOC object detection evaluation \cite{PASCAL-ijcv} to these new categories. We consider several variants of R-CNN on this metric and compare their performance. The dataset is available online (\url{http://www.cs.berkeley.edu/~gkioxari/}) and we hope that other researchers will use it to evaluate their methods.

This report is organized as follows.
In \secref{related} we present relevant past work. \secref{approach} describes our task-specific loss functions, while in \secref{experiments} we present in detail our experiments and results.

\section{Related work}
\seclabel{related}
While CNNs (\cite{lecun-89e}) achieved state-of-the-art performance on handwritten digit classification, most recognition work has used more traditional pipelines based on hand-engineered features. However, recently Krizhevsky \etal~\cite{krizhevsky2012imagenet} showed a big gain over existing methods in image classification on the ImageNet challenge~\cite{ILSVRC12}. This rekindled interest in CNNs. Donahue \etal~\cite{donahue2013decaf} found that Krizhevsky \etal's model can be used to extract features that are broadly useful for a large variety of tasks. Girshick \etal~\cite{girshick2014rcnn} found that using these features led to state-of-the-art object detection performance. They also found that ``fine-tuning'' the network (\ie, continued optimization of the network on a new task and/or dataset after starting from the initial ImageNet model) led to even greater gains. These results show that all these computer vision tasks are indeed related, and indicate that further improvements may be achieved through multitask training. CNNs have also given improvements on action classification and pose estimation, although gains have been somewhat smaller. Oquab \etal~\cite{Oquab14}  show small but significant improvements over state-of-the-art in action classification by finetuning a network pretrained on Imagenet. For pretraining, they add data from an additional 512 classes that they think are relevant. For pose estimation, Toshev \etal~\cite{toshev2014deep} learn a cascade of neural networks that regress to joint locations, with each successive layer refining the output of the last. At each stage, Toshev \etal augment data by randomly jittering input windows. The authors show improvements on many joints for multiple datasets. The gains mostly come in the low precision regime, indicating that the network does a good job of providing a rough location.

Previously leading, non neural network based approaches for the tasks of pose estimation used part detectors to provide accurate locations of the keypoints. Fischler and Elschlager \cite{Fischler73} introduced pictorial structure models in 1973. Felzenszwalb and Huttenlocher \cite{Felzenszwalb05} later rephrased PSM under a probabilistic framework. In their model, the body parts are represented as rectangular boxes and their geometric constraints are captured by tree-structured graphs. Ramanan \cite{parse2006}, Andriluka \etal \cite{Andriluka:2009} and Eichner \etal \cite{Eichner:2009} exploited appearance features to parse the human body. Johnson and Everingham \cite{Johnson:2010} replaced a single PSM with a mixture of PSMs in order to capture more variety in pose. Yang and Ramanan \cite{yang2012articulated} improved on pictorial structures mixtures by using a mixture of templates for each part. Sapp \etal \cite{Sapp:2010_2} enhanced the appearance models, including contour and segmentation cues. Wang \etal \cite{wang2011learning} used a hierarchy of poselets for human parsing. Pishchulin \etal \cite{Pishchulin2013cvpr} used PSM with poselets as parts for the task of human body pose estimation. Gkioxari \etal \cite{armlets2013} trained arm specific detectors and show state-of-the-art results on the PASCAL VOC dataset for arm pose prediction given ground truth location of the person, but did not propose an approach for whole-body pose estimation. Recently, Gkioxari \etal \cite{kposelets} introduced $k$-poselets, which are deformable part models composed of poselets, and showed state of the art performance on keypoint prediction for all keypoints on PASCAL VOC under a detection setting, where the location and number of the people in the image is unknown.

For the task of action classification from 2D images, most leading approaches are using part detectors to learn action specific classifiers. In detail, Maji \etal \cite{MajiActionCVPR11} train action specific poselets and for each instance create a poselet activation vector that is being classified using SVMs. Hoai \etal \cite{action} use body-part detectors to localize the human and align feature descriptors and achieve state-of-the-art results on the PASCAL VOC 2012 action classification challenge. Khosla \etal \cite{springer14_khosla} identify the image regions that distinguish different classes by sampling regions from a dense sampling space and using a random forest algorithm with discriminative classifiers.

\section{A Single Convolutional Neural Network for Multiple Tasks}
\seclabel{approach}

In this work, we use a single CNN that is trained jointly for multiple tasks. \figref{cnn} shows an example of a multitask CNN. Each task is associated with a loss function. We present loss functions for the tasks of person detection, pose estimation and action classification.

\noindent{\bf{Person detection.}}
Person detection is the task of predicting the location of people in an image. A detection is considered correct if the intersection over union of the predicted and ground truth bounding box  is more than a threshold (usually 0.5). During R-CNN fine-tuning for the task of person detection, a region $x$ is positive $(l=1)$ if it overlaps more than 0.5 with a ground truth person in the image, and negative $(l=0)$ if it overlaps less than 0.3. All the other regions are not considered. The output $y = [p_0, p_1]$ of the CNN is a two-dimensional probability vector, where $p_1$ indicates the probablity that $x$ is a person and $p_0 = 1-p_1$. The loss associated with the person detection task is a log loss
\begin{equation}
\loss_D = -(1-l) \log p_0 -l \log p_1
\eqlabel{det_loss}
\end{equation}

\noindent{\bf{Pose estimation.}}
Pose estimation is the task of predicting the location of specific keypoints in the human body. A prediction is correct if the estimation of the location of the keypoints is within some distance from the ground truth location. During R-CNN fine-tuning for the task of pose estimation, a region $x$ which overlaps more than 0.5 with an instance in the image is accompanied with a set of keypoint $(x,y)$ locations and visibility flags,  $\{(x_k,y_k,v_k)\}_{k=1}^{|K|}$, which belong to that instance. The keypoint locations are normalized with respect to the center and width and height of the region. The output of the R-CNN, $\{(\hat{x}_k,\hat{y}_k)\}_{k=1}^{|K|}$, is a set of predicted keypoint locations. The loss associated with the pose estimation task is the mean squared error between the prediction and the ground truth, ignoring keypoints which are not visible
\begin{equation}
\loss_P = \frac{1}{|K|} \sum_{k=1}^{|K|} v_k \left( (\hat{x}_k-x_k)^2 + (\hat{y}_k-y_k)^2 \right).
\eqlabel{pose_loss}
\end{equation}
For regions which do not overlap sufficiently with any ground truth instance, the loss is 0.

\noindent{\bf{Action classification.}}
Action classification is the task of predicting the action a person is performing. A prediction is correct if the predicted action is correct. During R-CNN fine-tuning for this task, a region $x$ is marked with an action label $l=\alpha$ if it overlaps more than 0.7 with a person performing action $\alpha$. We consider a higher threshold for the overlap compared to pose to ensure that the regions being used at training time are significantly close to the ground truth. The output of the R-CNN, $y = (p_1,...,p_{|A|})$, is a probability vector, where $p_\alpha$ indicates how likely $x$ contains action $\alpha$. The loss associated with this task is a log loss
\begin{equation}
\loss_A = -\sum_{\alpha=1}^{|A|} [l = \alpha] \log p_\alpha,
\eqlabel{action_loss}
\end{equation}
where $[P]$ is the Iverson bracket notation for a boolean statement $P$.
As for pose, regions which do not overlap sufficiently with ground truth always have a loss of 0.

If $x$ is an example with true labels $L_x=\{l_i\}_{i=1}^{N}$ and $f(x)=\{y_i\}_{i=1}^N$ is the output of the network, where $y_i$ is the output of the $i$-th task, the total loss of the network could be the weighted sum of the individual losses
\begin{equation}
\loss(f(x),L_x) = \sum_{i=1}^N \lambda_i \loss_i(y_i,l_i),
\eqlabel{sumloss}
\end{equation} 
where $\loss_i$ is the loss function associated with the $i$-th task and the parameter $\lambda_i$ is defined based on the importance of the task in the overall loss.

\begin{figure}[h]
\begin{center}
\includegraphics[height=3.5in]{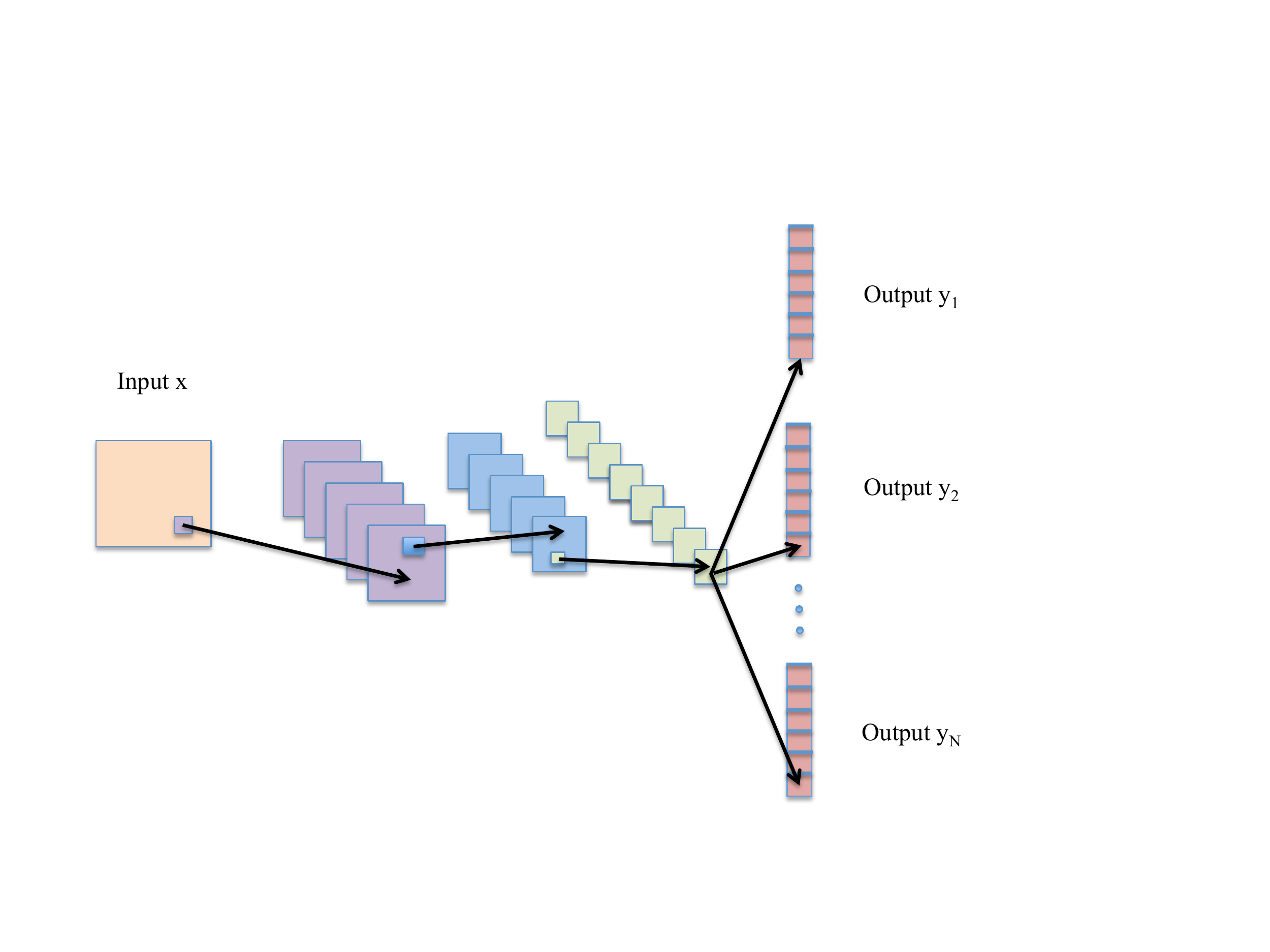}
\end{center}
\caption{An example of a convolutional neural network. The network takes a single input and computes N outputs. Each output is associated with a different loss function. The biggest part of the network is shared, while task specific parts exist only close to the output of the network.}
\figlabel{cnn}
\end{figure}

\section{Experiments}
\seclabel{experiments}
In this section, we describe in detail our experiments using CNNs for the tasks of pose estimation and action prediction. We use the CNN architecture, as defined by \cite{krizhevsky2012imagenet}, which has shown state of the art results for image classification on ImageNet. We build on the R-CNN work \cite{girshick2014rcnn} which recently demonstrated an impressive improvement from previous state of the art approaches on object detection on the PASCAL VOC and ImageNet.
We use region proposals generated by \cite{APBMM2014} and fine-tune the R-CNN starting from the ImageNet pretrained model on 1000 classes.

\subsection{Datasets \& Evaluation}


\noindent{\bf{Pose}}. For the pose estimation task we use the subset of PASCAL VOC 2012 detection dataset that has people, augmented with keypoint annotations. To train the R-CNN we use the \emph{train} set. For parameter selection and evaluation, we split the person images in the PASCAL VOC 2009 \emph{val} set into two sets: VAL09A is used for parameter selection, and VAL09B for test evaluation. 
The VAL09A subset consists of 723 images of 1529 instances and the VAL09B subset consists of 723 images of 1467 instances. 
For pose, VAL09A is used to train classifiers that estimate the confidence of the keypoint predictions produced by R-CNN. We evaluate the keypoint prediction using APK (average precision of keypoint), as defined in \cite{yang2012articulated}. APK measures the correctness of the keypoint predictions under a detection setting, where each prediction is associated with a score and those are subsequently used to compute precision-recall curves. A keypoint prediction is correct if the distance from the ground truth is less than a $0.2\cdot H$, where $H$ is the height of the torso of the corresponding ground truth instance.

\noindent{\bf{Action}}. For the action task we use the PASCAL VOC 2012 action dataset. To train the R-CNN we use the train set, which consists of 2296 images and 3134 instances (with their bounding box and action labels). There are in total 10 different actions. We use the PASCAL VOC 2012 test set to evaluate for the task of action classification. The action classification task assumes knowledge of the ground truth location of the people during test time. Therefore, not all person instances are necessarily annotated in the images. In order to evaluate for the task of action localization and classification, which we call \emph{action detection}, we annotated all the people on the validation set with bounding boxes and actions. This increased the number of annotated instances from 3144 to 5891. We use the augmented val set to evaluate for the task of action detection. For action classification we use the evaluation criterion defined by the PASCAL VOC action task, which computes the AP on the ground truth test boxes. For action detection, we measure the AP on all region proposals in the image, where a detection is correct if the overlap is more than 0.5 with a ground truth instance and the action is correctly predicted for that instance.

\subsection{Experimental Results}

\subsubsection{Pose Estimation}
We train a network, which we call \emph{Pose R-CNN}, with the loss function
\begin{equation}
\loss = \lambda_D \loss_D + \lambda_P \loss_P + \lambda_A \loss_A
\end{equation}
with $\lambda_D=\lambda_A=0$ and $\lambda_P=1$ (so, $\loss = \loss_P$).

During test time, for each input region the network predicts the locations of all the keypoints. In order to get a confidence score for those predictions, we train a linear SVM classifier for each of the keypoints. The SVM classifier is trained using as positive examples the fc7 features of the regions that make a correct prediction for that keypoint (as defined in the APK metric), and as negative all other regions. We train $|K|$ such classifiers on the VAL09A set, where $|K|$ is the number of keypoints. For each of the $|K|$ keypoints, we measure the AP of that keypoint on VAL09B. \tableref{apk} shows the results of Pose R-CNN. For comparison with the state of the art approach, we show the perfomance of $k$-poselets \cite{kposelets} on the same set of images. Note that Pose R-CNN outperforms $k$-poselets with a relative improvement of 19.7\%.

\begin{table*}[t!]
\centering
\renewcommand{\arraystretch}{1.2}
\renewcommand{\tabcolsep}{1.2mm}
\resizebox{\linewidth}{!}{
\begin{tabular}{@{}l|r*{12}{c}|cc@{}}
\textbf{APK [$\alpha=0.2$]}   & Nose  & R\_Shoulder   & R\_Elbow   & R\_Wrist     & L\_Shoulder  & L\_Elbow  & L\_Wrist  & R\_Hip & R\_Knee & R\_Ankle & L\_Hip & L\_Knee & L\_Ankle  & \emph{mAP}  
\\
\hline
$k$-poselets   & 42.9 & 27.1 & 12.2 & \phz3.4 & 27.3 & 11.8 & \phz2.8 & \bf{10.6} & \bf{\phz4.4} & \phz3.8 & \bf{11.4} & \bf{\phz4.9} & \phz3.2 & 12.7 \\
\hline
\textbf{Pose R-CNN} & \bf{52.0} & \bf{32.5} & \bf{16.6} & \bf{\phz5.9} & \bf{32.1} & \bf{14.6} & \bf{\phz5.6} & \phz9.7 & \phz4.0 & \bf{\phz4.6} & 10.8 & \phz4.8 & \bf{\phz4.8} & \bf{15.2}
\end{tabular}
}
\vspace{0.1em}
\caption{APK average precision (\%) on the second half of PASCAL VOC val 2009 person dataset (VAL09B). We report keypoint prediction AP for $k$-poselets \cite{kposelets}. Pose R-CNN is better for almost all keypoints.}
\tablelabel{apk}
\vspace{-0.5em}
\end{table*}

\figref{apk_res} shows examples of the predictions of Pose R-CNN on ground truth regions of VAL09B. (We only use ground truth regions for visualization purposes.)

\begin{figure*}[t!]
  \centering
  \includegraphics[height=0.09\textheight]{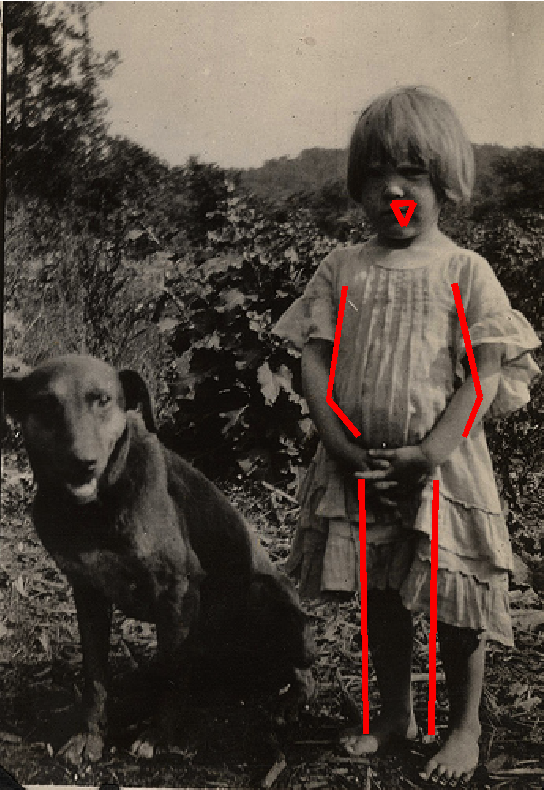}
  \includegraphics[height=0.09\textheight]{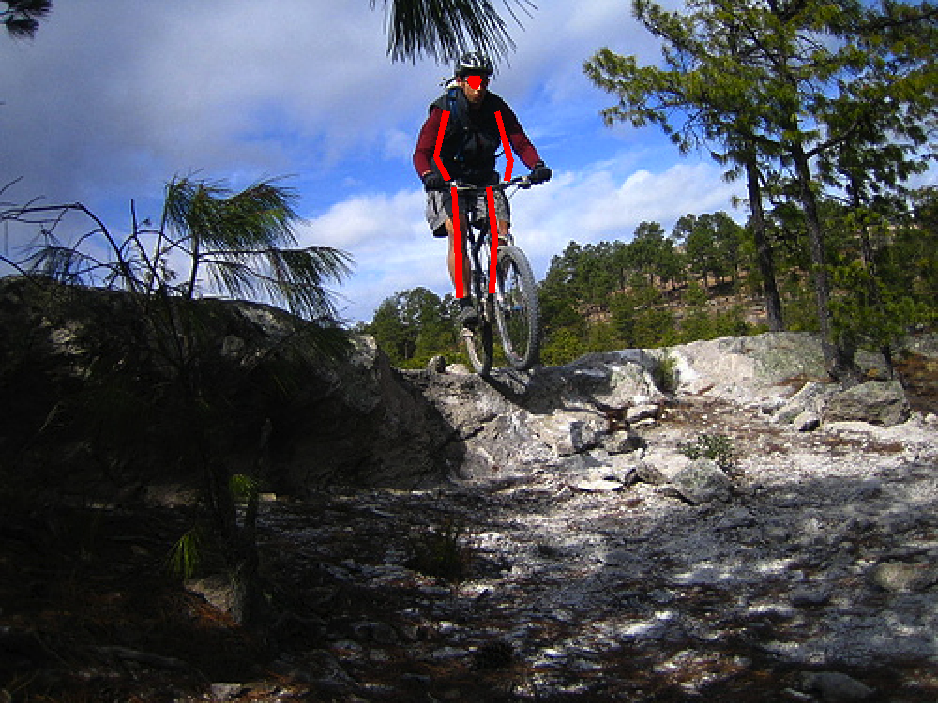}
  \includegraphics[height=0.09\textheight]{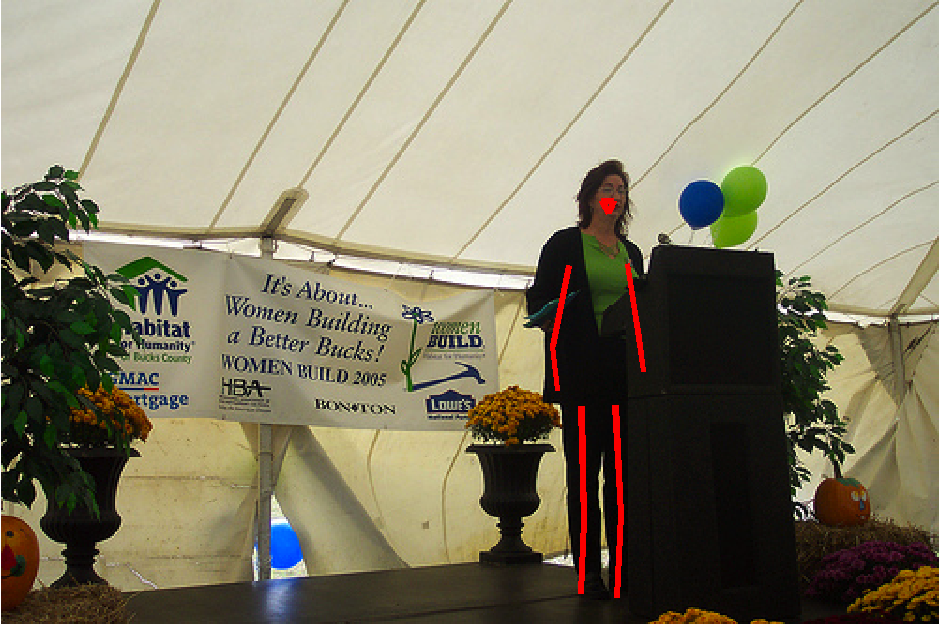}
  \includegraphics[height=0.09\textheight]{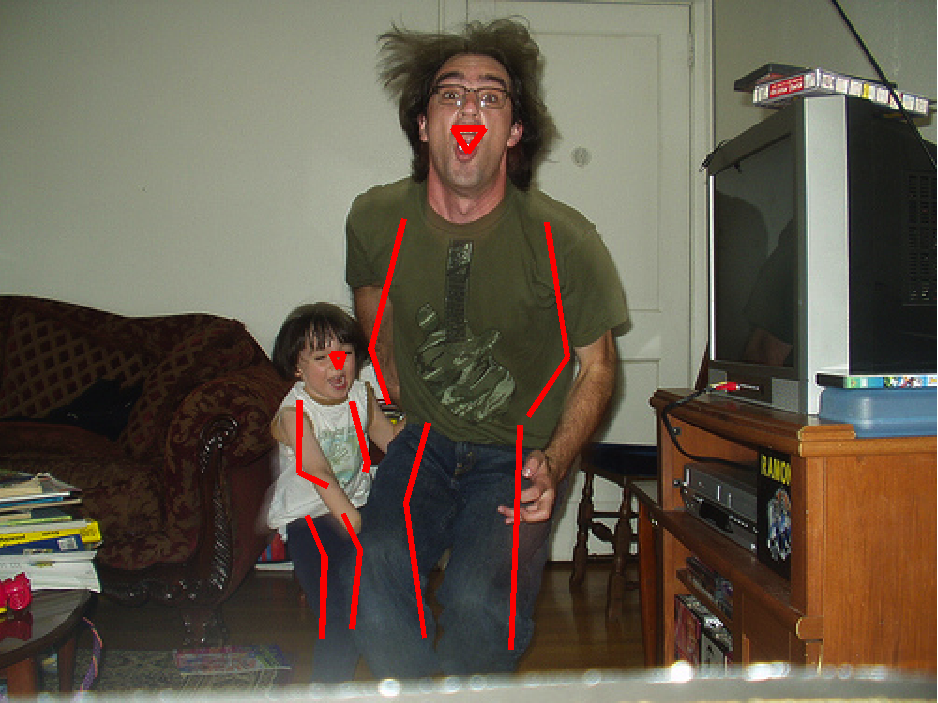}
  \includegraphics[height=0.09\textheight]{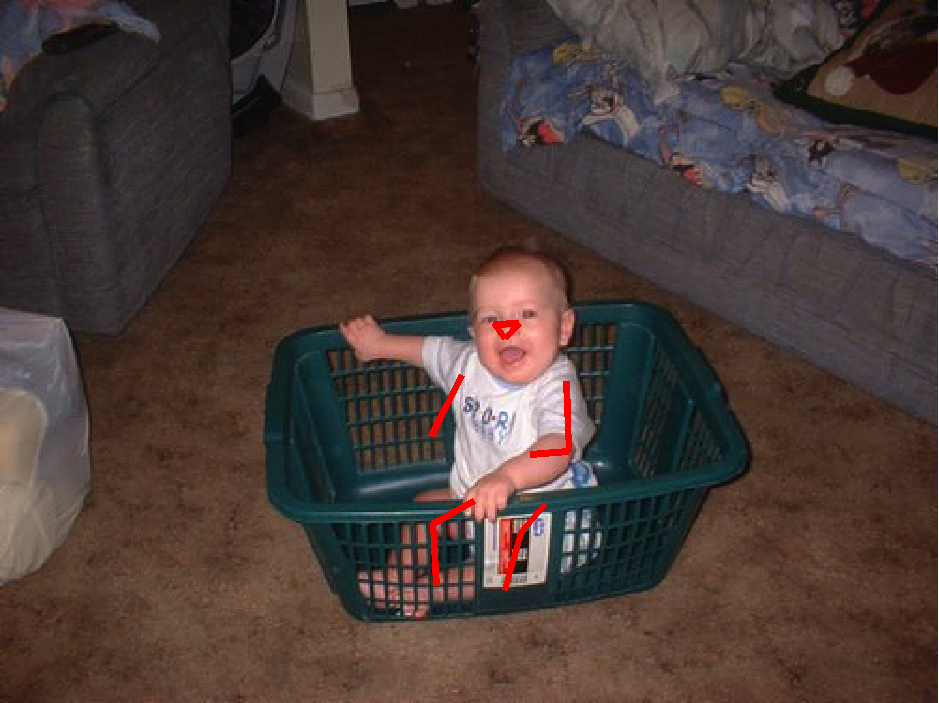}
  \includegraphics[height=0.09\textheight]{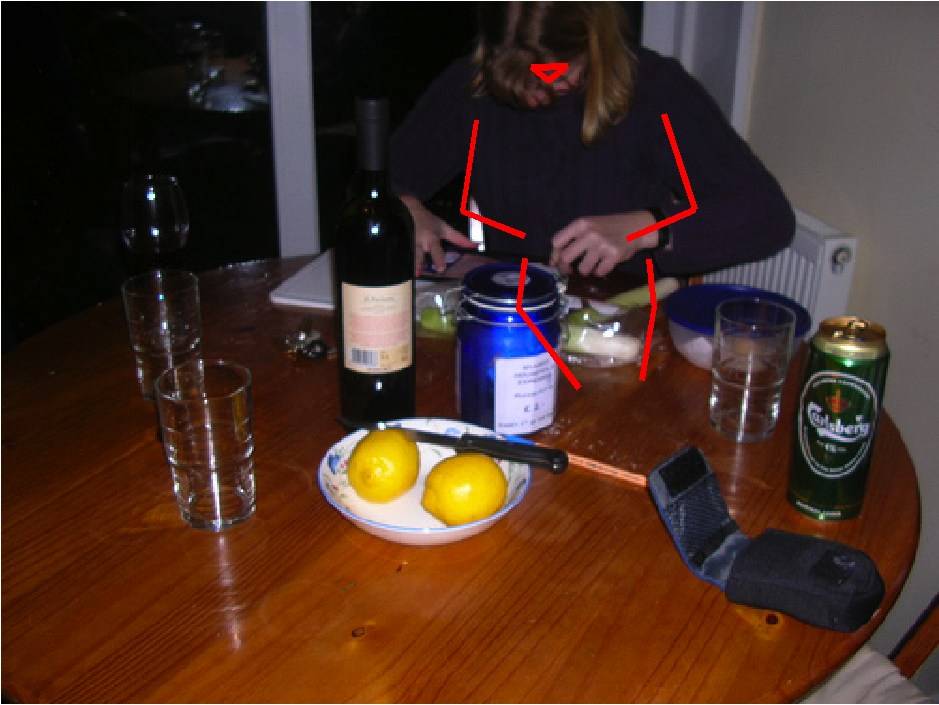}
  \includegraphics[height=0.09\textheight]{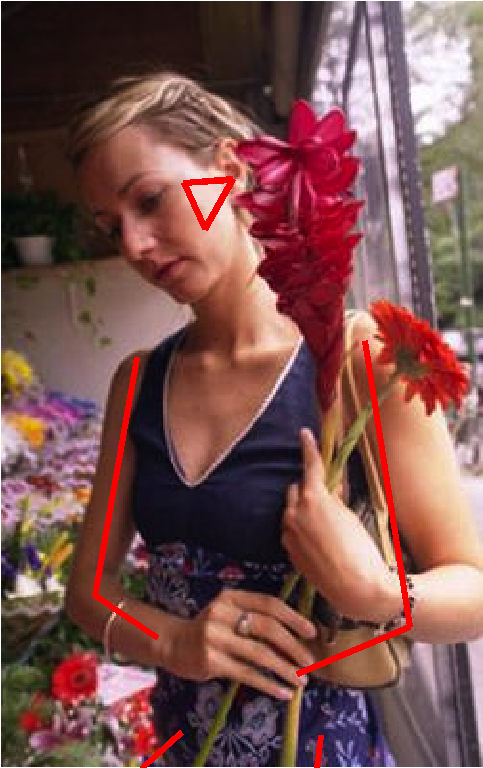}
  \includegraphics[height=0.09\textheight]{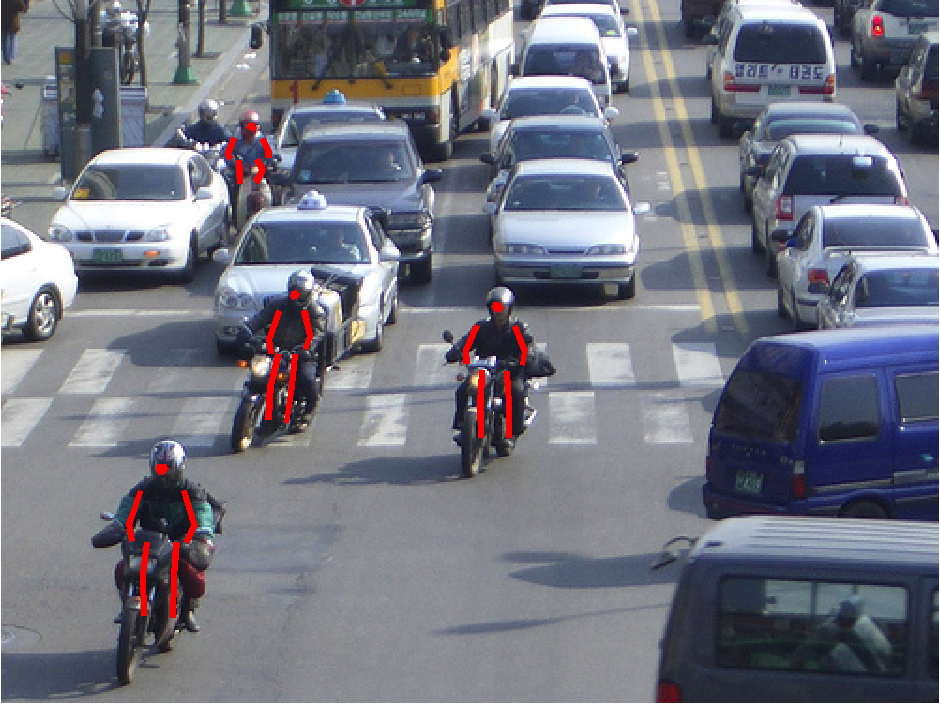}
  \includegraphics[height=0.09\textheight]{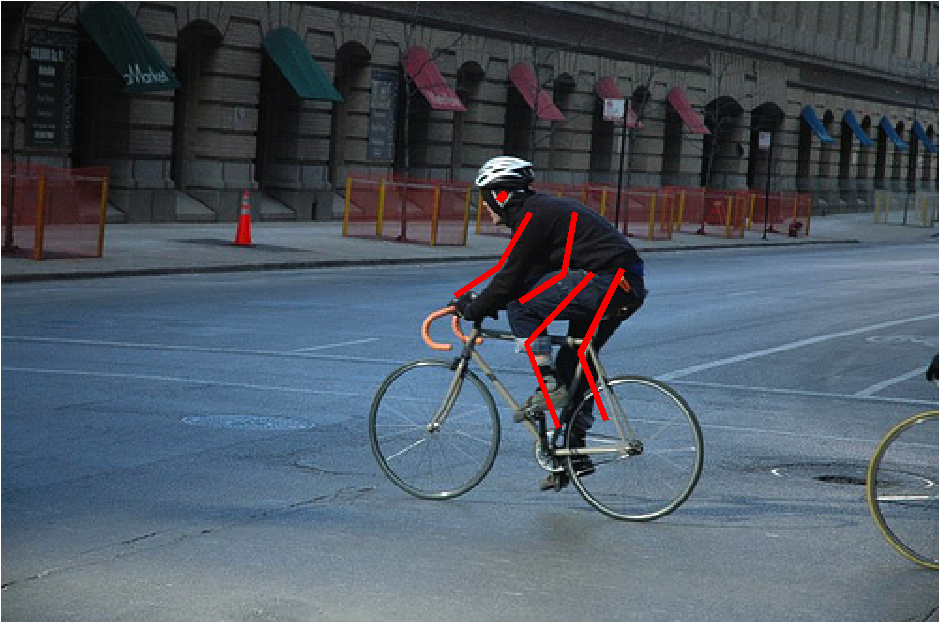}
  \includegraphics[height=0.09\textheight]{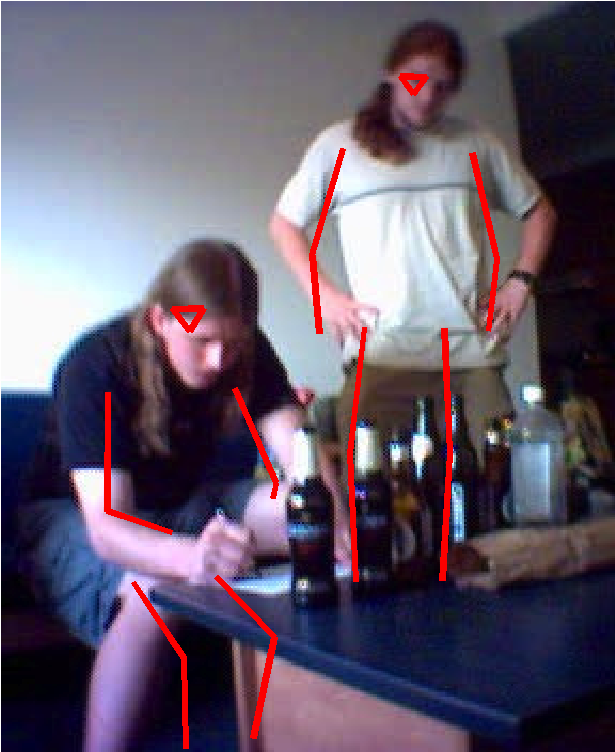}
  \caption{Examples of keypoint predictions from Pose R-CNN on ground truth regions on VAL09B. Ground truth information is only used for visualizatin purposes. The Nose and Eyes predictions are connected in a triangle, while the keypoints of the arms and the legs are visualized using stick figures.}
  \figlabel{apk_res}
\end{figure*}

\subsubsection{Action Classification}
We train a network, which we call \emph{Action R-CNN}, with the loss function
\begin{equation}
\loss = \lambda_D \loss_D + \lambda_P \loss_P + \lambda_A \loss_A
\end{equation}
with $\lambda_D=\lambda_P=0$ and $\lambda_A=1$ (so, $\loss = \loss_A$).

For the task of action classification, the location of the people performing the action is considered known. Since there is too little data to train a large CNN, we augment the training set with regions that overlap more than 0.7 with the ground truth regions. This leads to an increase of 100x of the training set. 

During test time, a ground truth region is being marked with a score for each action. This score can come directly from the softmax output of the CNN or by training action specific classifiers (linear SVMs) on top of fc6 or fc7 features. In practice, we observed that fc6 features work slightly better than the softmax output and fc7, perhaps due to overfitting of the network. In addition, in order to make use of context, which is very useful for such classification tasks, we rescore our predictions. In particular, we train linear SVM classifiers on feature vectors composed of: (1) the predicted score of the action from Action R-CNN, (2) the maximum action scores from Action R-CNN of all the other instances in the image, if any, and (3) the maximum scores of the objects \{horse, bike, motorcycle, tv monitor\} that overlap more than 0.1 with the ground truth region. The score of the objects were computed using a regular R-CNN trained on the 20 object categories of the PASCAL VOC detection challenge, similar to \cite{girshick2014rcnn}.

\tableref{action_classif} shows the results of Action R-CNN. We also show leading state of the art approaches in this field, based on the PASCAL VOC 2012 leaderboard \cite{action}. Note that \cite{Oquab14} use a CNN, like us, but fine-tune their network by learning two additional fully connected layers. They get leverage from the objects involved and the context around the people performing the action by initializing their network with a model trained on ImageNet with 1512 categories, including horse, bike etc., and by using bigger regions centered at the ground truth location. Action R-CNN perfoms slightly better on average compared to the rest of the approaches.

\begin{table*}[t!]
\centering
\renewcommand{\arraystretch}{1.2}
\renewcommand{\tabcolsep}{1.2mm}
\resizebox{\linewidth}{!}{
\begin{tabular}{@{}l|c*{9}{c}|cc@{}}
\textbf{AP}   & Jumping & Phoning & Playing Instrument & Reading & Riding Bike & Riding Horse & Running  & Taking Photo & Using Computer & Walking & \emph{mAP}  
\\
\hline
Stanford \cite{action} & 75.7  & 44.8 & 66.6 & \bf{44.4} & 93.2 & 94.2 & 87.6 & 38.4 & \bf{70.6} & \bf{75.6} & 69.1  \\

Oxford \cite{action} & \bf{77.0} & \bf{50.4} & 65.3 & 39.5 & 94.1 & \bf{95.9} & \bf{87.7} & 42.7 & 68.6 & 74.5 & 69.6      \\

Oquab \etal \cite{Oquab14} & 74.8 & 46.0 & 75.6 & 45.3 & 93.5 & 95.0 & 86.5 & 49.3 & 66.7 & 69.5 & 70.2 \\
\hline
Action R-CNN + fc6 SVM& 76.2 & 47.4 & \bf{77.5} & 42.2 & \bf{94.9} & 94.3 & 87.0 & \bf{52.9} & 66.5 & 66.5 & \bf{70.5}

\
\end{tabular}
}
\vspace{0.1em}
\caption{AP (\%) on Test 2012 for action classification. We report numbers for leading approaches in the task, according to the PASCAL VOC leaderboard \cite{action}. Action R-CNN is better on average.}
\tablelabel{action_classif}
\end{table*}

\subsubsection{Action Detection}
The task of action classification assumes knowledge of the location of the people. This makes the task a lot easier, since it skips the difficult step of detection. A real application cannot assume perfect localization of the objects to be classified. On the other hand, action detection is the task of localizing and classifying the people as perfoming one of $|A|$ actions. A correct prediction is one that overlaps more than 0.5 with a ground truth instance and predicts the correct action label. 

We train a network, which we call \emph{Detection-Action R-CNN}, with the loss function
\begin{equation}
\loss = \lambda_D \loss_D + \lambda_P \loss_P + \lambda_A \loss_A
\end{equation}
with $\lambda_P=0$ and $\lambda_D=\lambda_A=1$.

We use the PASCAL VOC 2012 detection and action train set. We use the softmax output of the network to make action predictions, since it proved to work better than fc6 or fc7 features with SVMs. \tableref{action_recog} shows the perfomance of Action R-CNN and Detection-Action R-CNN, as well as the performance of \emph{Detection R-CNN} trained solely for the task of detection ($\lambda_D=1,\lambda_P=\lambda_A=0$). The Detection R-CNN and Action R-CNN make action predictions by using action specific SVM classifiers, which are trained on fc6 features on the PASCAL VOC action train set, after cross validating across choices of features.

\begin{table*}[t!]
\centering
\renewcommand{\arraystretch}{1.2}
\renewcommand{\tabcolsep}{1.2mm}
\resizebox{\linewidth}{!}{
\begin{tabular}{@{}l|c*{9}{c}|cc@{}}
\textbf{AP}   & Jumping & Phoning & Playing Instrument & Reading & Riding Bike & Riding Horse & Running  & Taking Photo & Using Computer & Walking & \emph{mAP}  
\\
\hline
Action R-CNN + fc6 SVM & 14.4 & 8.4 & 9.4 & 4.7 & 9.6 & 19.0 & 16.4 & 10.8 & 3.1 & 3.7 &  10.0 \\

Detection R-CNN + fc6 SVM & 19.5 & 9.1 & 11.9 & 7.5 & 9.9 & 20.1 & 24.4 & 5.3 & 1.8 & 9.3 & 11.9   \\

\hline
\bf{Detection-Action R-CNN} & \bf{29.6} & \bf{22.4} & \bf{28.4} & \bf{16.8} & \bf{26.2} & \bf{35.2} & \bf{28.1} & \bf{22.7} & \bf{15.8} & \bf{20.6} & \bf{24.6}

\
\end{tabular}
}
\vspace{0.1em}
\caption{AP (\%) on the augmented action val 2012 dataset. We report numbers of various types of R-CNN networks. The network trained jointly for detection and action (Detection-Action R-CNN) outperfoms the rest of the R-CNNs by a significant margin.}
\tablelabel{action_recog}
\end{table*}

\figref{action_res} shows examples of the Detection-Action R-CNN. For every image, we show the top 3 activations along with their predicted action labels and scores. It is clear from the examples that the network prefers regions around the face to predict \emph{phoning} and \emph{taking photo}, while it prefers regions that include the whole person for actions such as \emph{walking} and \emph{jumping}. Also, activations for \emph{playing instrument}, \emph{riding bike} and \emph{riding horse} seem to be driven by the presence of objects in the proximity of the region.

\begin{figure*}[t!]
  \centering
  \includegraphics[height=0.09\textheight]{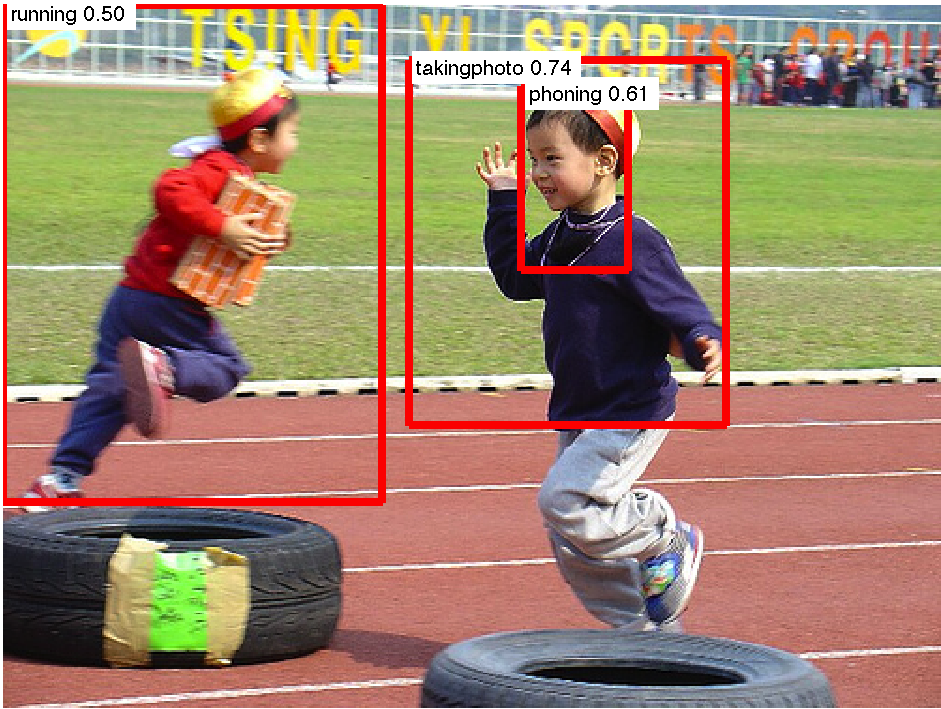}
  \includegraphics[height=0.09\textheight]{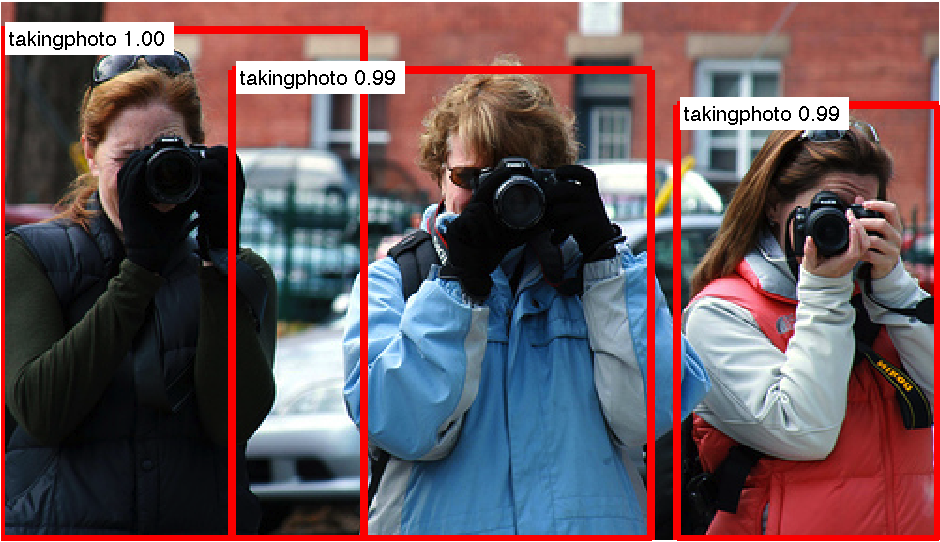}
  \includegraphics[height=0.09\textheight]{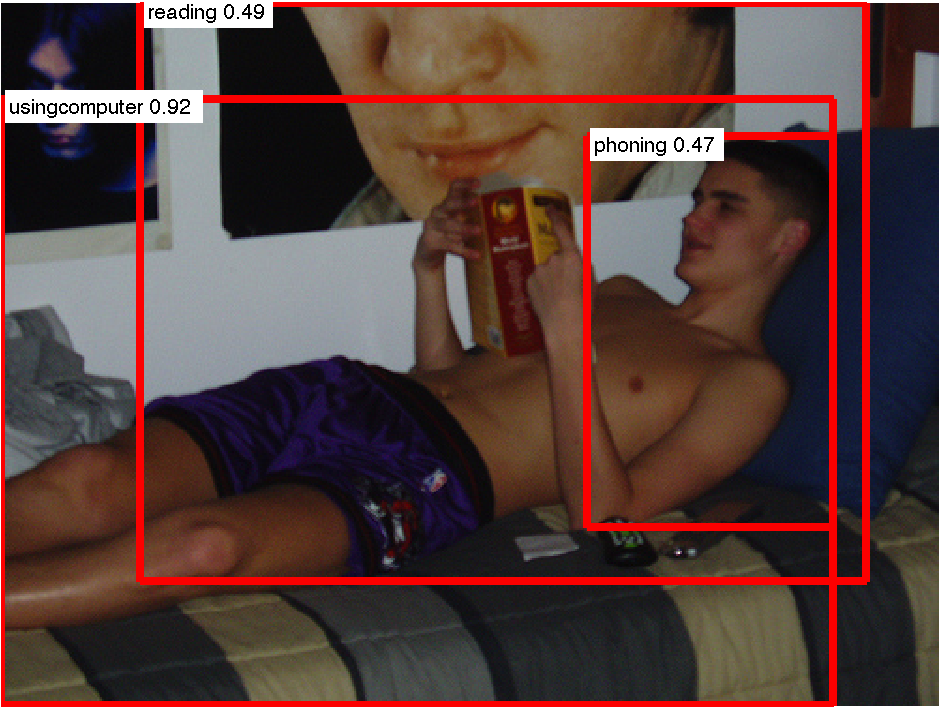}
  \includegraphics[height=0.09\textheight]{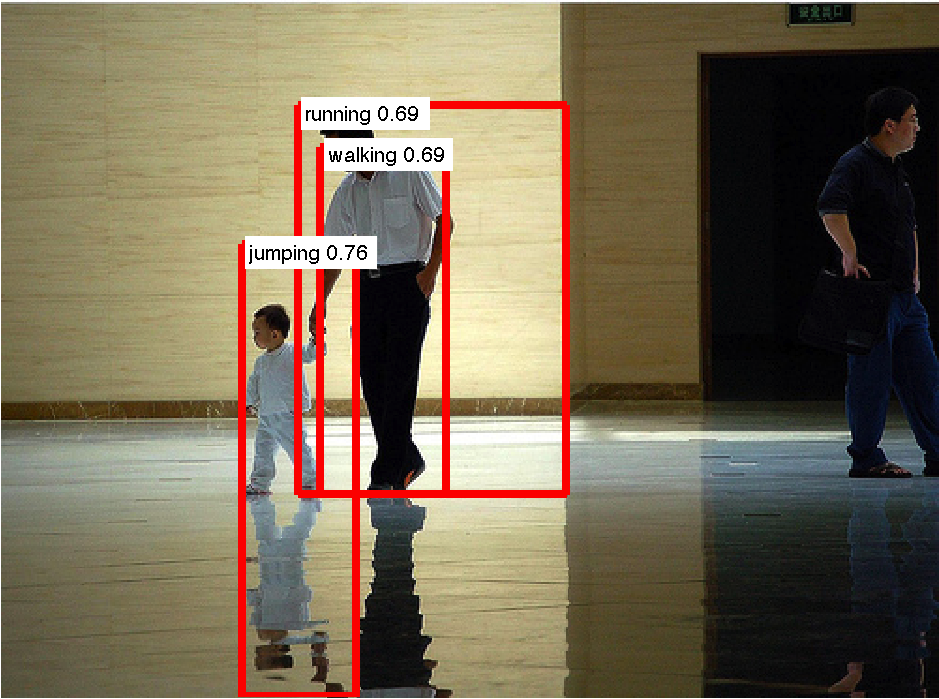}
  \includegraphics[height=0.09\textheight]{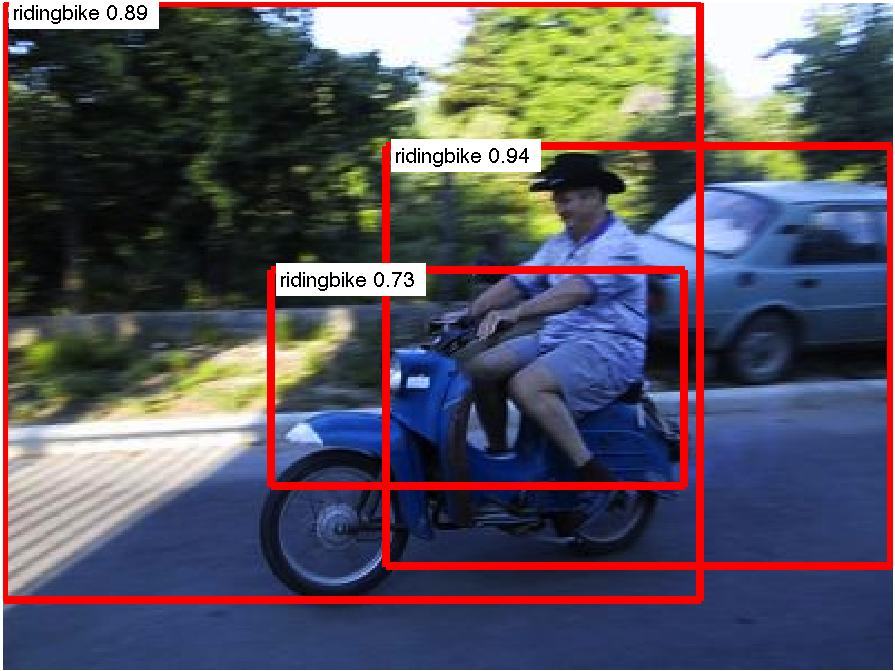}
  \includegraphics[height=0.09\textheight]{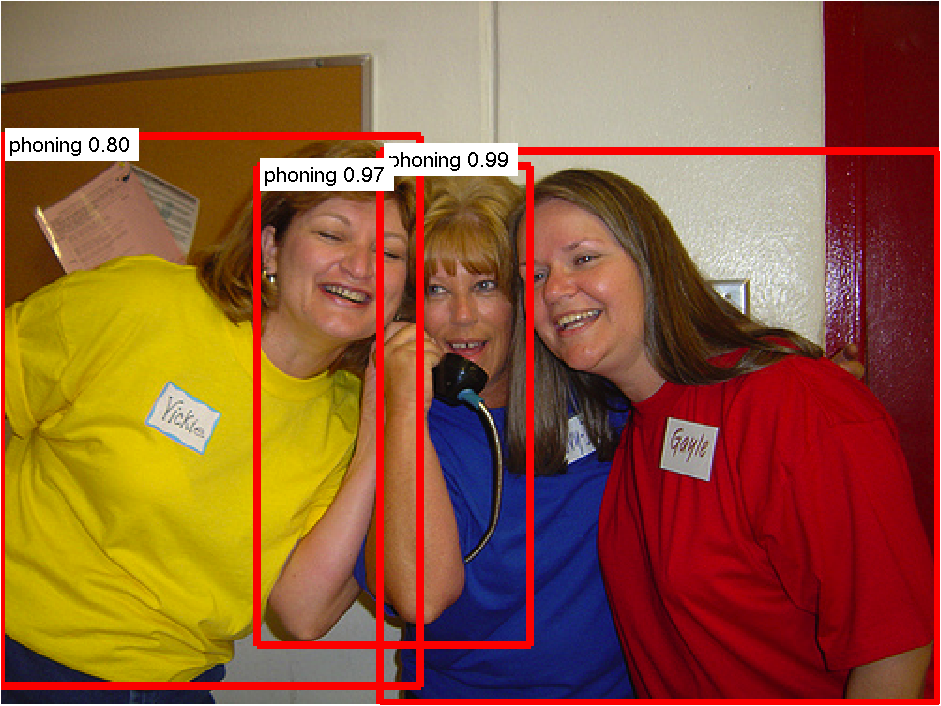}
  \includegraphics[height=0.09\textheight]{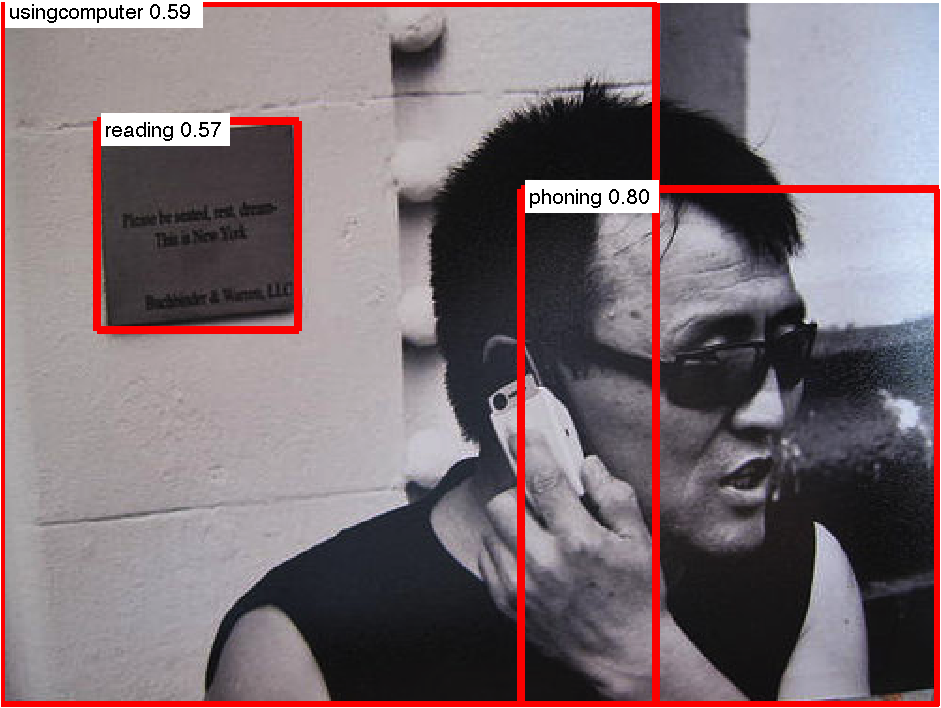}
  \includegraphics[height=0.09\textheight]{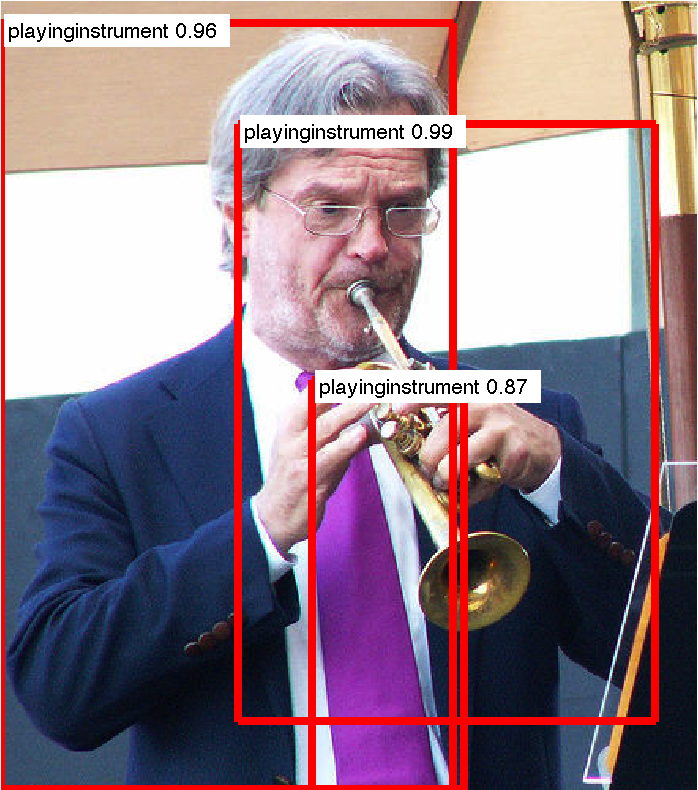}
  \includegraphics[height=0.09\textheight]{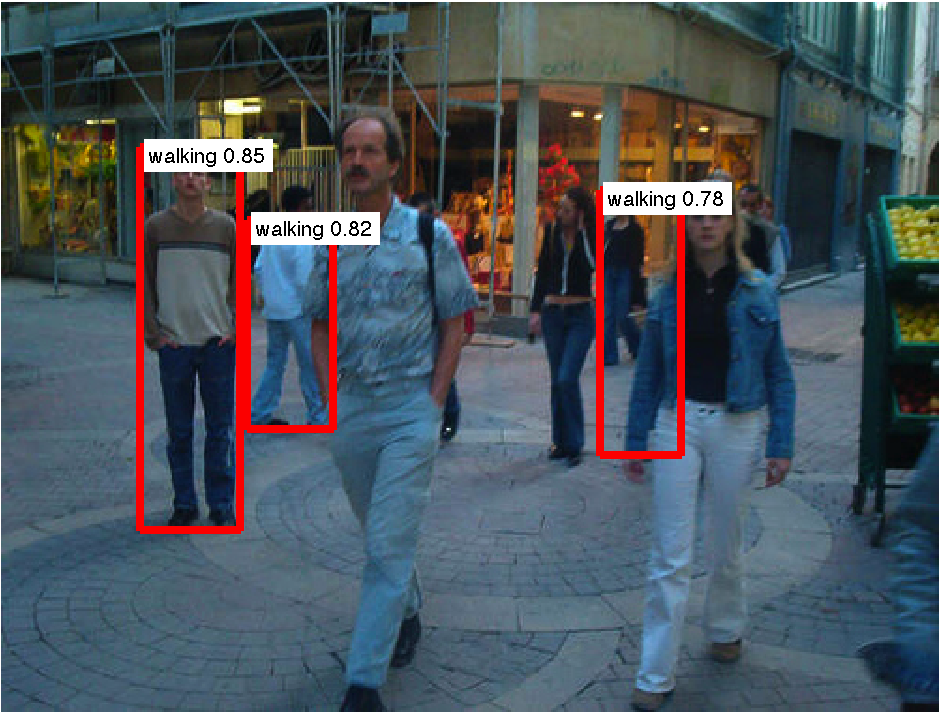}
  \includegraphics[height=0.09\textheight]{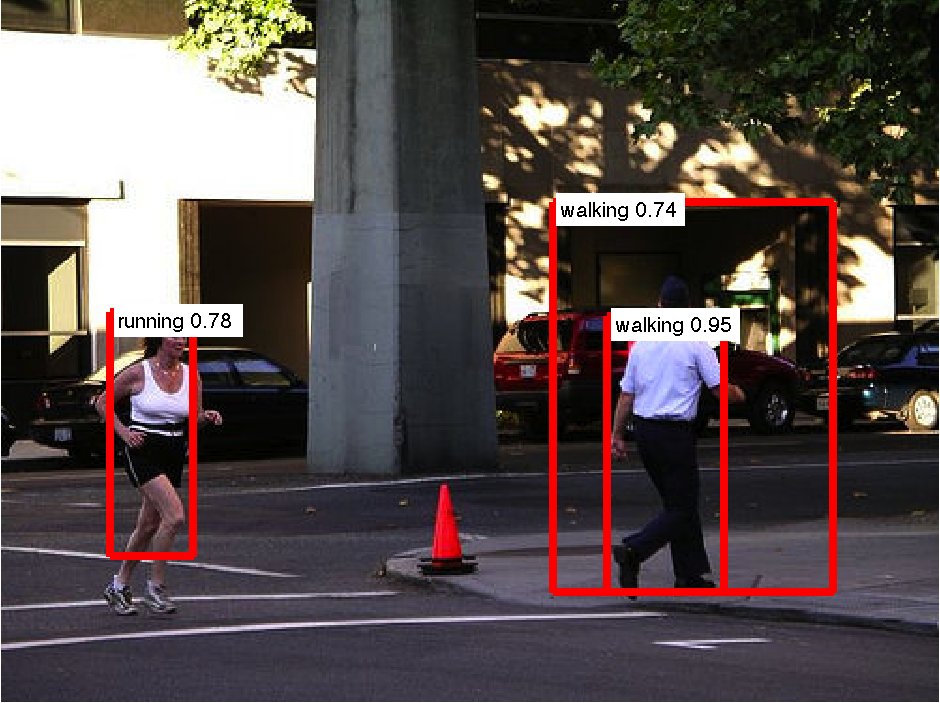}
  \includegraphics[height=0.09\textheight]{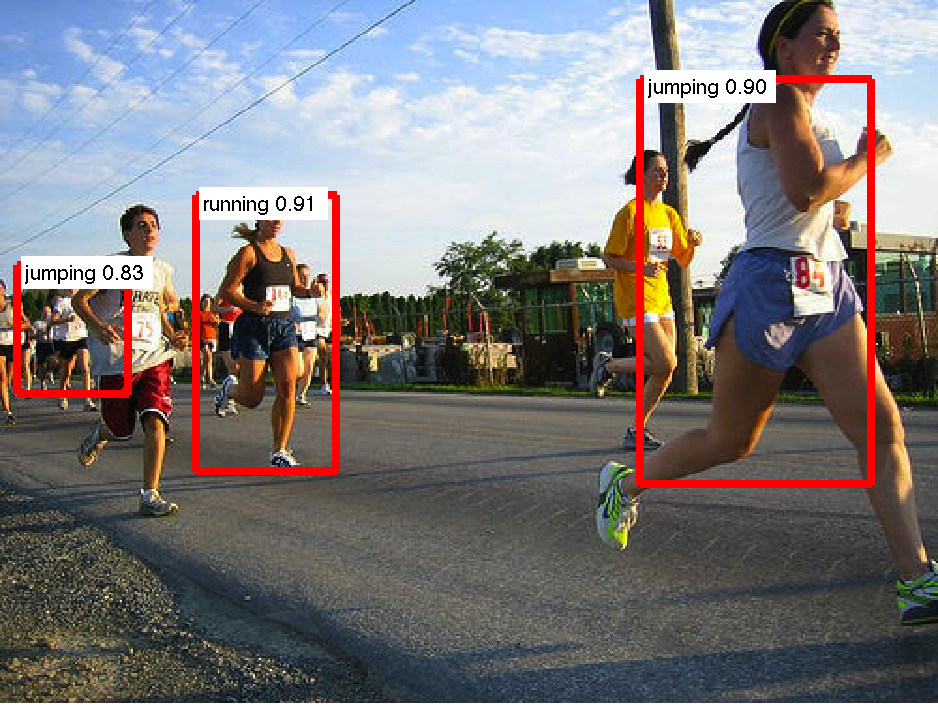}
  \includegraphics[height=0.09\textheight]{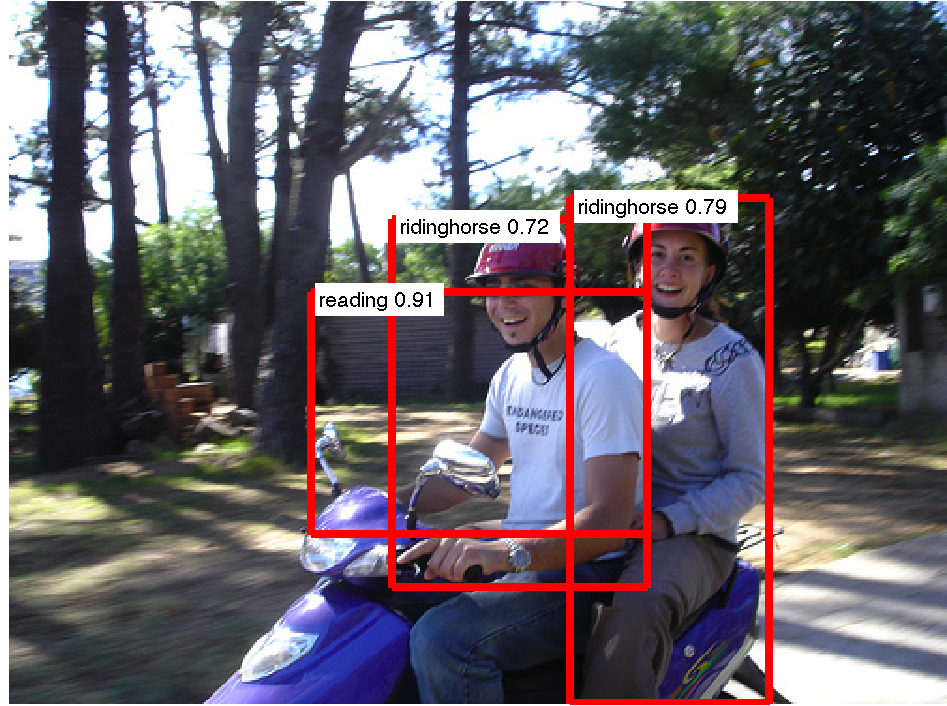}
  \includegraphics[height=0.09\textheight]{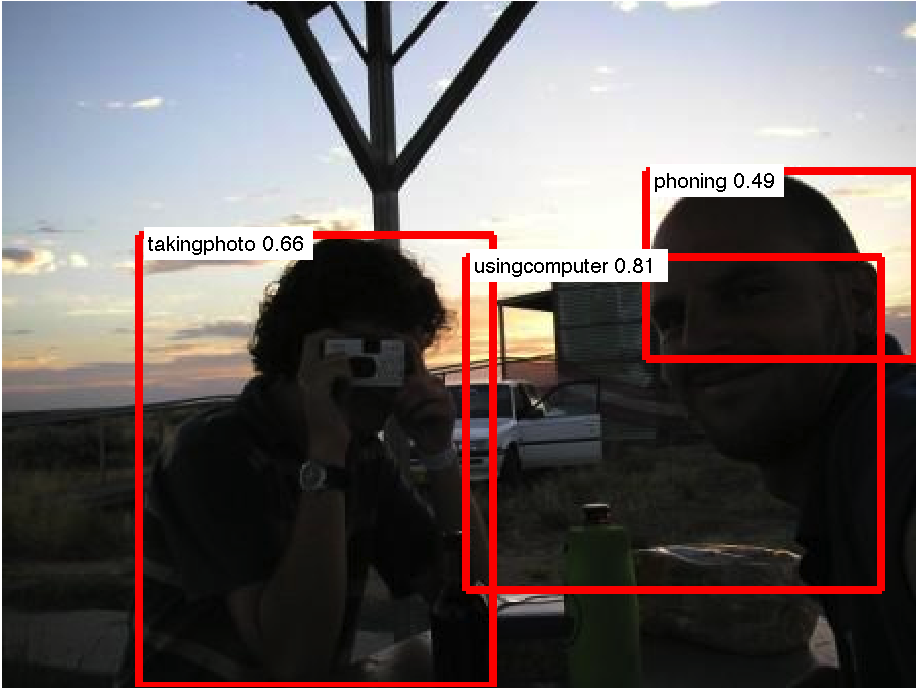}
  \caption{Examples of action detections from the Detection-Action R-CNN on the PASCAL VOC 2012 action val set. (We encourage the reader to view this figure on a computer and zoom in.)}
  \figlabel{action_res}
\end{figure*}

\subsubsection{Person Detection}
For the task of person detection, the Detection R-CNN with a SVM classifier on fc7 features achieves an AP of 54.8\% on the PASCAL VOC 2012 detection test set. (Note that the original work of R-CNN  \cite{girshick2014rcnn} reports 53.2\% with no bbox regression. The difference in performance is probably due to the different nature of the regions \cite{APBMM2014}.) The performance of the Detection-Action R-CNN with a similar SVM classifier achieves 56.0\% on the PASCAL VOC 2012 test set for the task of person detection. This suggests that networks that are jointly trained for detection and other tasks might be useful for the task of detection as well.

\subsubsection{A single network for Detection, Pose and Action}

We train a network for all three tasks jointly, which we call \emph{Detection-Pose-Action} R-CNN, with loss function
\begin{equation}
loss = \lambda_D loss_D + \lambda_P loss_P + \lambda_A loss_A
\end{equation}
where $\lambda_D=\lambda_P=1$ and $\lambda_A=2$. We choose a higher value for action classification to make sure that the task has a significant contribution to the total loss, since there is significantly fewer training data for action compared to detection and pose.

The joint network for the three tasks performs on average similar to the networks trained for specific tasks individually. Specifically, for the task of person detection, the joint network achieves 56.4\% on the PASCAL VOC 2012 test set. For pose, the mean AP for keypoint prediction on the VAL09B set is 15.5\%, while for action detection, the mean AP is 21.6\%.

The joint network behaves very similar to the individual networks, while at the same time it is $N$-times faster. It involves finetuning a single network instead of $N$ networks and processing the evaluation data once instead of $N$ times.

\section*{Acknowledgments}
This work was supported by the Intel Visual Computing Center, ONR SMARTS MURI N000140911051, ONR MURI N000141010933, a Google Research Grant and a Microsoft Research fellowship. The GPUs used in this research were generously donated by the NVIDIA Corporation.

{\small
\bibliographystyle{ieee}
\bibliography{nips2014}
}

\end{document}